# Mitigating Adversarial Attack for Compute-in-Memory Accelerator Utilizing On-chip Finetune

Shanshi Huang, Hongwu Jiang and Shimeng Yu
School of Electrical and Computer Engineering, Georgia Institute of Technology
Atlanta, GA, USA
Email: shimeng.yu@ece.gatech.edu

*Abstract*—Compute-in-memory (CIM) has been proposed to accelerate the convolution neural network (CNN) computation by implementing parallel multiply and accumulation in analog domain. However, the subsequent processing is still preferred to be performed in digital domain. This makes the analog to digital converter (ADC) critical in CIM architectures. One drawback is the ADC error introduced by process variation. While research efforts are being made to improve ADC design to reduce the offset, we find that the accuracy loss introduced by the ADC error could be recovered by model weight finetune. In addition to compensate ADC offset, on-chip weight finetune could be leveraged to provide additional protection for adversarial attack that aims to fool the inference engine with manipulated input samples. Our evaluation results show that by adapting the model weights to the specific ADC offset pattern to each chip, the transferability of the adversarial attack is suppressed. For a chip being attacked by the C&W method, the classification for CIFAR-10 dataset will drop to almost 0%. However, when applying the similarly generated adversarial examples to other chips, the accuracy could still maintain more than 62% and 85% accuracy for VGG-8 and DenseNet-40, respectively.

## I. INTRODUCTION

Though deep neural networks (DNNs) have yielded outstanding results in a variety of applications, including speech recognition, image classification, and natural language processing [1], there is a growing concern regarding adversarial attack which aims to fool the model with manipulated input samples (e.g. adding with noises [2]). The prior works on adversarial attack and defense mostly were performed from the software's perspective [3], there are rarely any discussions from the hardware's perspective. In this work, we will explore the adversarial attack and defense on the actual inference chip based on the compute-in-memory accelerator, which is becoming attractive for power-constrained edge intelligence platform [4].

As DNNs are generally data and compute intensive, frequent data movements between logic and memory units limit the energy efficiency on traditional Von Neumann architecture. In recent years, there are increasingly efforts on developing specific hardware accelerators to run large-scale DNN models from the cloud to the edge. For example, systolic architecture such as TPU [5] employs many digital multiply and accumulate (MAC) engines close to a large global buffer (i.e., SRAM) to reduce the cost of data movement. As a more aggressive approach, compute-in-memory (CIM) architecture [4] merges the computation directly into the memory sub-arrays that ideally addresses the memory-wall problem. The weights of a DNN model could be mapped as the conductance of the memory cells in the sub-array, while the input vector is loaded in parallel as the voltage to the rows, then the multiplication is done in analog fashion, and the current summation along columns represents weighted sum. In principle, CIM could be implemented by different device technologies. SRAM with modified bit-cell and array periphery could enable parallel access as demonstrated in recent silicon prototype chips [6]. Emerging non-volatile memory (eNVM) technologies also provide promising solutions due to a smaller cell size and potential of multi-bit per cell, yielding a higher integration density at the same technology node [7]. Besides, because of the non-volatile nature and near-zero leakage, the eNVM-based CIM is more attractive to edge devices. No matter which kind of memory technologies is used, ADC is commonly essential as an important part of periphery circuitry to convert the analog partial sum back to digital signal for further processing. In other words, CIM is essentially mixed-signal compute, thus the variations are unavoidable. As reported in prior work [8], inference accuracy measured in CIM prototypes generally is degraded from the software baseline. The primary variation sources include the cell-to-cell variation for eNVMs and the intrinsic ADC offset. Cell-to-cell variation could be minimized by iterative write-verify technique with tolerable overhead for inference engine [9]. A more critical challenge is the intrinsic ADC offset introduced by the manufacturing process variation. As a result, the ADC offset may noticeably degrade the inference accuracy and cause different chip instances having different inference results even for the same input. It is noted that when ADC offset introduces quantization error because of the process variation, these offset patterns are static once the chip is fabricated.

As mentioned earlier, it has been suggested that DNN is under the threat of adversarial examples, which could fool the network easily while will not affect human's decision. In general, adversarial attacks could be categorized into white-box attack and black-box attack based on the information of target model exposed to the adversary. For the white-box attack [10,11], the adversary has full access to the DNN model architecture and weights. Whereas, only external access to the network (e.g., input and output) could be used for the black-box attacks [12,13]. The white-box attack can often achieve higher attack success rates compared to the black-box attack [10]. While the white-box attack will cause more serious problem, it is not easy for the adversary to get the access to a private model in cloud. However, for the edge device, it is physically accessible by anyone and thus could leak the model information at high risk. Although defense methods have been proposed for the white-box attack, the extra algorithmic calculation will introduce speed and power overhead. This is undesired for the edge device which has limited power budget and demands real-time response.

In this work, we leverage the ADC offset pattern (which is believed to be detrimental to the inference accuracy) but finetune the model weights to take its advantage against the adversarial attack on the CIM accelerator. In our evaluation, we find that the accuracy drop could be compensated by finetuning DNN parameters that adapt to ADC offset. This finetune, while recovering the inference accuracy, makes the DNN parameters slightly different from chip to chip, which brings us a byproduct: the chip will be robust to the adversarial examples generated by attacking other chips or software baseline. Explicitly, even if the adversary attacks one chip instance by manipulating the adversarial input, he/she could not use the same adversarial input to attack all the other chip instances due to the uniqueness of the DNN model for each chip.

This paper is organized as follows. Section II introduce the principle of CIM scheme, and backgrounds of adversarial attack and defense. Section III presents our methodology including ADC offset modeling and hybrid on-chip/off-chip finetune procedure. Section IV shows our evaluation setup and evaluation results of the proposed defense method to the white-box attack. Section V concludes the paper.

## II. BACKGROUND

### A. Pricniple of CIM

Convolutional neural network (CNN) is a major class of DNNs. Convolution is an important mathematical operation widely used in image processing which extracts the features of images by filters. The inference operation of convolution is essentially the vector-matrix multiplication (VMM). Sometimes, the depth of the input channel and output channel could be very large that make it hard to fit weights of one layer into a single memory array considering slow access and excessive energy consumption. Array partitioning [14] can be introduced to parallelize the computation into multiple sub-arrays. To maximize the input data reuse, a novel mapping method was proposed in [15]. In this method, the weights at different spatial location of each kernel correspond to different sub-kernels. These sub-kernels are mapped into different subarrays. Because of the window sliding manner of convolution, sub-kernels of each position will see the input to their neighbor at the window sliding direction. In this case, by passing the used input vectors in the same direction as the kernel "slides over" the input tensor, the input vectors can be reused among the subarrays efficiently.

The crossbar nature of memory array is a natural substrate for implementing VMM in a highly parallel manner. As shown in Fig. 1, the crossbar array consists of perpendicular rows and columns with the memory cell located at each cross-point. Weights in the filters are mapped as the content of the cells. The VMM operation is performed as follows: read voltages representing the input feature map are applied to all the rows so that the read voltages are multiplied by the memory cells at each cross-point. The current through each device is summed up along columns. Different columns represent filters for different output channels, who should see the same input thus all the columns work at the same time in parallel. Typically, ADCs are needed at the end of the column to convert the analog current to the digital output so that the subsequent processing such as activation and pooling could be performed in the digital domain. In principle, VMM could be done in fully parallel fashion if asserting all the rows and all the columns simultaneously. In practice, multiple rows/columns could be partially turned on due to the sensing resolution of ADCs or the mismatch of column pitch to the peripheral circuitry's dimension.

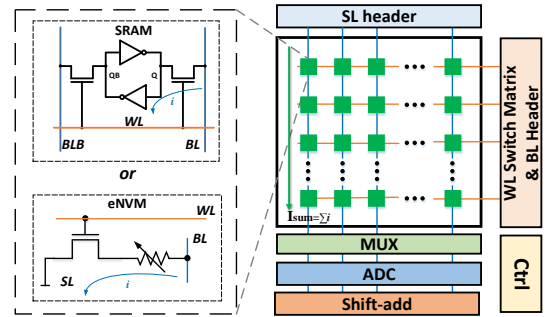

**Fig. 1 The crossbar array for CIM architecture with SRAM or eNVM.**

### B. Adversarial attack

While AI applications become more prevailing in our daily life, their security vulnerability becomes a serious concern. On one side, the DNN model becomes a valuable asset since it may bring financial profit. The model stealing will ease the life of adversary to get a well-trained model or private training dataset information. On the other side, the DNN model itself could be adversarial attacked, poison attacked, etc., thereby being hampered from the intended usage.

Poisoning attack is an attack mechanism that happens during the training of the machine learning model. Generally, it is done by injecting "bad" data into the training dataset so that the model achieves good performance during training but becomes more fragile during inference [16]. Software techniques [17] are also proposed to defend model from the poisoning attack to some extent.

The adversarial attack has been widely studied in the software domain, especially on DNN-based classification tasks. Adversarial examples, which could be defined as "inputs formed by applying small but intentionally worst-case perturbations to examples from the dataset, such that the perturbed input results in the model outputting an incorrect answer with high confidence" [2], could hurt the functionality of DNN models while cause no uncertainty on human's decision. In physical worlds, camera noise, stick on the target object, etc. may lead to adversarial examples causing safety-critical situations like self-driving car. In general, the adversarial attacks could be divided into two categories, which are white-box attack and black-box attack, according to the exposure level of network information to the adversary.

*1) White box attack*

For white box attack, the attackers are supposed to have the access to the network structure and parameters, thus make it easy to generate adversarial examples by utilizing gradient [2], optimization [11], etc.

*2) Black box Attacks*

In the black-box attack, the adversary has no access to the model parameters. What he/she could only do is to feed the input to the network and observe the output from the network. So, in principle, black-box try to "extract" network information from the input-output pairs. A typical method used by black-box attack is to train a new model using the new dataset generated from the victim model, which is called substitute model. The adversary has full access to the substitute model and thus could apply the white-box attack on it [18].

*C. Adversarial defense*

In order to protect the integrity of DNN models, different countermeasures have been proposed against adversarial examples. Since most white-box attacks strongly depend on the gradient to generate adversarial examples, one straight way is to make gradient inaccessible. For example, in [19,20], some non-smooth or non-differentiable functions are added to preprocess the inputs so that it will not cause problem when calculating weights' gradient during training but cut the gradient to the input. As mentioned in [2], it claimed that the adversarial examples exist since the data space explored by the network is limited. Thus, a simple but efficient way to improve the robustness is to retrain the network with adversarial examples included with correct labels, which is called adversarial training. Since the adversarial examples are generated by adding some noise to move the input to some "untouched" region, one could force it back by input preprocessing such as de-noise [21] and compression [22]. In addition, since the attack is normally applied on a certain model, some randomness can be introduced into the network parameters so that the adversary could not predict which exact set of parameters of the model will be used for classification [23].

In this work, instead of trying to defend against adversarial examples of a certain model in software, we aim to reduce the transferability of the adversarial examples among actual chips. It works like the aforementioned software defense method of introducing randomness into the network parameters. For CIM architecture that employs ADCs, there are intrinsic process variations that will introduce quantization error. As shown in previous prototype chip measurement results, even with precisely designed ADCs, the accuracy will be low if the ADCs all share the same references [24]. To achieve high accuracy, the references of each ADC need to be adjusted independently. This requirement for ADC reference adjustment and independent reference for each ADC brings additional hardware overhead. Alternatively, we could adjust the weights with several retrain epoch, namely "finetune" process. When the model is adapted to the ADC offset pattern, the inference accuracy could be recovered. The overhead of model finetune in software is much less than the implementation of adjustable ADC references on-chip. Now we could take advantages of the model finetune to reduce the transferability of the adversarial examples from software baseline to actual chips, and from one chip to another.

III. METHODOLOGY

*A. ADC offset modeling*

As aforementioned, ADC plays an important role to support mixed-signal processing and has a significant impact on the inference accuracy. As reported in prior CIM design, ADC quantization will lead to a degraded accuracy performance [8]. There are mainly two causes: partial sum quantization loss and ADC offset. The resolution of ADC is determined by the partial sum precision required from the sub-array, which is then affected by the number of rows that are turned on simultaneously and the memory cell precision. When a large weight matrix is partitioned into several sub-arrays, partial sums are obtained from multiple sub-arrays and then accumulated. Considering huge hardware overhead introduced by ADCs, The resolution of ADC is usually less than full precision of the partial sum, causing quantization loss. The quantization loss of partial sums will be also gathered while accumulating partial sums, which leads to more accuracy loss. The impact of quantization loss could be relieved by sacrificing hardware performance (e.g. turn on less number of rows simultaneously) or reducing required input/weight precision (e.g. network compression). On the other hand, ADC offset caused by process variation is a more critical issue. Unlike cell conductance variation which could be manually tightened by aggressive write-verify [9], ADC offset for a given circuit topology is purely defined by manufacturing process variation. The intrinsic ADC offset makes the partial sum read-out from one memory sub-array different from the correct value after quantization. Despite offset could be compensated by advanced offset cancellation techniques but with significant area overhead [25], the tight column pitch of CIM subarrays limits using advanced ADCs in order to preserve the parallelism of the computation.

The popular ADC topologies in prior CIM works are Flash-ADC and successive-approximation-register (SAR)-ADC due to their simplicity and suitability. Flash-ADC is made of cascading comparators. For an N-bit converter, the circuit employs $2^N - 1$ comparators. After comparison, the thermometer code is then encoded to the binary code. Oppositely, SAR-ADC only hires a single comparator but needs several iterative cycles to finish bit-by-bit comparison based on binary search algorithm. Both ADC designs are investigated to evaluate our proposed adversarial defense scheme. In both traditional memory and CIM designs, sense amplifier (SA) is typically employed as comparator as it can achieve high speed at low power consumption. In our evaluation, we use a simple current-mode SA design as shown in Fig. 2 (a), which is based on the classic cross-coupled latch structure (P1-P2, N1-N2). This SA operates in 2 phases: precharge & sensing. During the precharge phase, the PRE signal goes low and the bit-lines are precharged to VDD. During the sensing phase, the PRE and SAEN signals are both high which activates the cross-coupled structure to pull the outputs to the appropriate full logic level. In CIM design, bitline current becomes smaller as load resistance is larger, corresponding to smaller partial sum value.

As a case study, we build a 5-bit ADC with such SA design to explore the offset impact on ADC outputs. Fig. 2 (b) shows the relationship between ADC pass rate and partial sum value according to the 1,000 Monte Carlo simulation runs with TSMC

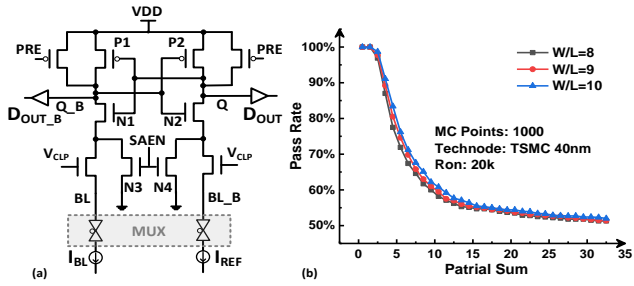
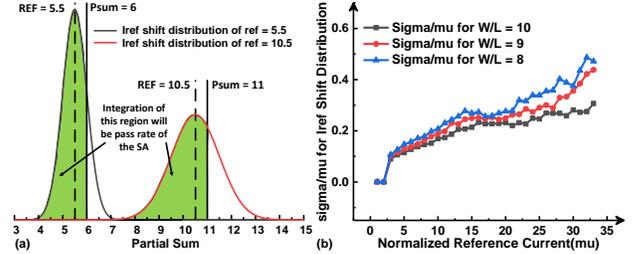

**Fig. 2 (a) Latch-based current-mode SA. (b) Sense pass rate for 5-bit ADC.**

**Fig. 3 (a) Sense pass rate to Iref offset conversion. (b) Sigma/mu of the Gaussian distribution of Iref offset converted from sense pass rate.**

40nm PDK which offers eNVM process [26]. Here the sense pass rate is defined as the probability to sense correctly between the bitline current (partial sum) and its nearest reference current (Iref). As shown in the plot, the sense pass rate decreases with increasing partial sum value (namely increasing bitline current). This observation was also reported in the silicon data [7] and could be explained as follows: the absolute sensing voltage difference is determined by bitline current and reference current. When the bitline current becomes larger as partial sum increases, the sensing voltage variation caused by ADC offset (especially N3-N4) tends to dominate, thus the output is more decided by the intrinsic SA variation. Another reason is that column current is inversely proportional to bitline resistance. This property leads to smaller ADC step size for higher ADC level (partial sum), in which the variation causes higher error rate. We also find that the pass rate slightly increases as transistor W/L ratio increases.

As shown in Fig. 3 (a), the pass rate of SA is converted as the cumulative probability of Iref being smaller than the Psum, corresponding to the green shade area. Assuming the Iref distribution follows the Gaussian function, we regard ADC offset as the distance between shifted Iref and its ideal value and the sigma of Iref could be back-calculated. As abovementioned, process variation is caused by manufacturing, thus the Iref shift is a static variation over time. From the DNN model's standpoint, the Iref shift could be regarded as partial sum quantization bias. For 5-bit Flash-ADC, since each SA performs the comparison at its certain reference level (31 different levels), each SA has different shift value. Oppositely, SAR-ADC only employs the same SA during the entire comparison, thus the Iref should be shifted to the same direction for each reference level. Based on the observation that the sigma over mu ratio of the shift distribution is increased with Iref (Fig. 3(b)), we scale up the absolute shift value (normalized to Iref) by the ratio. For Flash-ADC, as random shifts from different SA are independent, offset compensation is possible. However, the mechanism of SAR-ADC determines that offset of different levels shifts towards one direction. As a result, while sensing the same partial sum, SAR-ADC has bigger offset than Flash-ADC as shown in Fig.4. Fig. 4 (a) (b) shows the output distribution with offset for Flash-ADC while Fig. 4 (c) (d) shows the output distribution for SAR-ADC. From the plot, we can also observe that ADC with smaller transistor size (e.g. W/L) could induce more offset, resulting in lower sense pass rate. As we utilize the process variations in this work, SAR-ADC with larger variation is a better choice.

### B. Procedure of weights finetune

We now discuss the on-chip/off-chip hybrid finetune procedure to mitigate adversarial attack from the hardware standpoint. During the retraining, the feedforward propagation (inference) is first performed on-chip, and then the backpropagation and weight update are done off-chip by software. The detailed process is as follows: we will run the inference on a specific chip that captures its specific ADC offset pattern, then the prediction of the inference will be compared with the ideal label for the loss function; after obtaining the estimated loss, weights are updated through backpropagation in software; finally, the memory cells will be reprogramed to the new weights possibly with write-verify. In our evaluation, the partial sum will sample the ADC offset from the estimated distribution as in Fig. 3(a), and the distorted partial sum will be used as output feature map and saved for error and gradient calculation. The backpropagation and weight update are all directly use floating-point calculation as done in software.

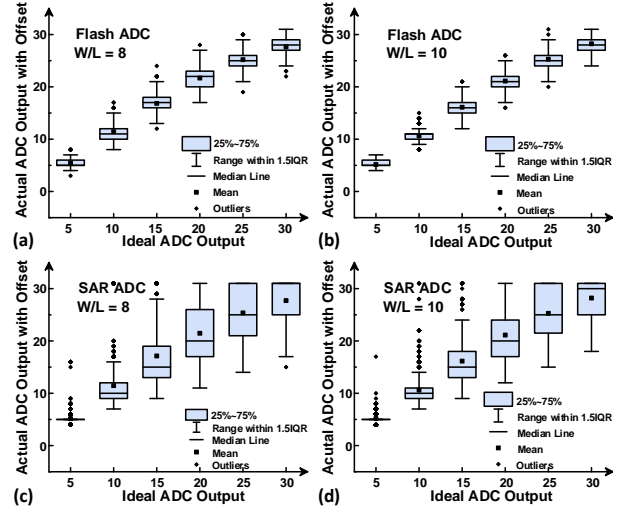

**Fig. 4 Simulated ADC output with offset sampled from the Iref distribution**

## IV. EVALUATION RESULTS

We evaluate the proposed hybrid finetune defense method with VGG-8 and DenseNet-40 networks for CIFAR-10 dataset. The precision setting is 8-bit activation and 2-bit weight for VGG-8, and 8-bit activation and 8-bit weight for DenseNet-40. The software baseline accuracy is ~92% for both networks. For the weight finetune process, the batch size of retrain is 200, which means there are 250 iterations to finish the finetune in one

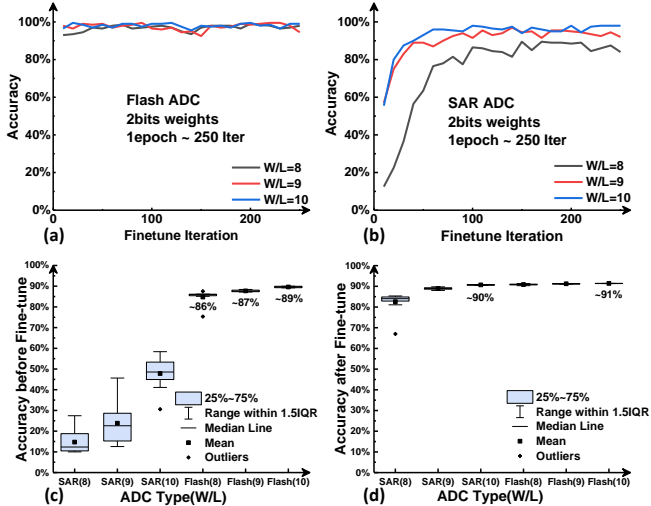

**Fig. 5 (a)** Retrain curve of Flash-ADC in one epoch. **(b)** Retrain curve of SAR-ADC in one epoch. **(c)** Accuracy distribution before finetune with ADC offset on VGG-8. **(d)** Accuracy distribution after finetune with ADC offset on VGG-8.

epoch. Fig. 5 (a) & (b) shows the retrain curve of one specific chip that implements VGG-8 which uses Flash-ADC and SAR-ADC, respectively. Chip with either Flash-ADC or SAR-ADC could recover the accuracy, however Flash-ADC has less initial accuracy drop and easier to be retrained to recover the high accuracy. This is consistent with the analysis in Section III on the possible compensation of SA offsets for Flash-ADC. It is also seen that as the W/L decreases, it will be more difficult to retrain the model to recover the accuracy under process variations. When the W/L is small, which means the sense pass rate is also low, the accuracy may could not be fully recovered. It needed to be pointed out that the W/L reported here appears high since we use the minimum length ($L_{min}$) as the L in our simulation. Generally, two to three times of $L_{min}$ will be used in the analog circuit to avoid very large process variation. We did not optimize the ADC with advanced offset cancellation techniques. Here our goal is just to show that by changing W/L, we could achieve different degrees of the process variation. Fig. 5 (c) & (d) presents the accuracy distribution of several retrain tests before finetune and after finetune collected from multiple chips, respectively. From the plot, we could observe that the accuracy recovery from finetune is generally achievable.

We evaluate a white-box attack called Carlini and Wagner (C&W) Attack [11] on VGG-8 and Desnet-40 model in three different cases: Attack original model; Attack retrained digital model; Attack retrained chip (as described in Fig. 6). Table 1 presents accuracy performance under three attack cases with different ADC settings (ADC variation decreases from A to D). As shown in table, chip accuracy could be generally recovered to baseline accuracy (above 90%) by retrain. While applying the adversarial attack to the original software baseline model (model0), accuracy drops to ~0% which means the attack is effective. However, when the generated adversarial examples from the software baseline are applied to chip1 (attack case 1), retrained on-chip network could still preserve relatively high accuracy (~75% for VGG-8, ~84% for DenseNet-40).

**Network Finetune:**
- Train a network without variation, saved as model0
- Load model0 to chip1 which has a set of ADC variation specified for it. Fine-tune the network to recover the accuracy.
- Load model0 to chip2 which has a set of ADC variation specified for it. Fine-tune the network to recover the accuracy.

**Case1: Attack original model:**
- Attack model0, which is the pure digital network, to generate a set of images: *adversarial examples*
- Apply *adversarial examples* to chip1

**Case2: Attack retrained digital model:**
- Read the digital weights on chip1 out and load it to network in pure digital version. In this case the digital model knows nothing about the adc offset and thus will experience performance degradation, we call this as model1.
- Attack model1, which is the pure digital network, to generate a set of images: *adversarial examples*
- Apply *adversarial examples* to chip1

**Case3: Attack retrained chip:**
- Attack chip2, which is a hybrid process that inference is performed on chip and backpropagation is calculated by software, to generate a set of images: *adversarial examples*
- Apply *adversarial examples* to chip1

**Fig. 6** Pseudo-code of adversarial attack methods on actual chips.

For attack case 2, the adversary first read out the retrained model (model1) which is finetuned to fit the ADC variation on chip1 and uses it to generate adversarial examples in software environment. The accuracy of model1 has already degraded before attack due to the lack of considering the ADC offsets according to variation level and its digital accuracy drops to ~0% after attack. However, while applying generated adversarial examples on chip1, accuracy is still relatively high thanks to the ADC variation. We could observe that the higher the ADC variation is, the more robust the chip will be to the adversarial examples from the digital model.

For attack case 3, instead of generating adversarial examples fully in software, adversarial examples are obtained in hybrid manner that inference is performed on chip2 and backpropagation is calculated by software. Under such attack, chip2 itself fails to generate even one correct answer but chip1 can still maintain a certain accuracy since the chip variation is distinct from one to another and thus make the models unique for each chip. Besides, compared to VGG-8, DenseNet-40 is more robust to transferred adversarial examples (~86% compared to 64%) as shown in Table 1.

In Table 2, we varied the distance matric used in C&W attack on VGG-8 ($L_0, L_2, L_\infty$), and we could see that the proposed defense is effective regardless of the used norm type. It should be pointed out that preserving CIFAR-10 accuracy ~60% under adversarial attack is comparable with the prior software defense techniques [27].

## V. CONCLUSION

In this paper, the threats of adversarial attacks on CIM-based machine learning edge inference engine are identified. We first explore ADC offset modeling in CIM designs and proposed an on-chip finetune scheme against adversarial examples. Our evaluation results show that by utilizing the ADC offset, the

Table 1: Accuracy performance under C&W attack ($L_2$)

| Chip config. | Chip Information | | | Attack original model | | Attack retrained digital model | | | Attack retrained chip | |
|---|---|---|---|---|---|---|---|---|---|---|
| | ADC type | W/L | Retrained accuracy | Software Attack(model0) | Attack on chip1 | Digital accuracy (model1) | Software Attack(model1) | Attack On chip1 | Chip2 acc. after attack | Attack on chip1 |
| VGG-8 | | | | | | | | | | |
| A | SAR | 9 | 89.39% | 0.61% | 73.95% | 74.75% | 0.09% | 83.43% | 0% | 62.10% |
| B | SAR | 10 | 90.87% | | 75.12% | 83.89% | 0.24% | 78.78% | | 64.80% |
| C | Flash | 9 | 91.36% | | 74.10% | 89.31% | 0.15% | 65.73% | | 65.10% |
| D | Flash | 10 | 91.46% | | 74.40% | 90.54% | 0.21% | 51.22% | | 64.30% |
| DenseNet-40(k=24) | | | | | | | | | | |
| A | SAR | 9 | 91.04% | 0% | 84.59% | 20.04% | 0% | 87.69% | 0% | 87.20% |
| B | SAR | 10 | 91.52% | | 83.11% | 35.25% | 0% | 89.52% | | 85.25% |
| C | Flash | 9 | 91.50% | | 85.56% | 62.71% | 0% | 87.62% | | 86.80% |
| D | Flash | 10 | 91.81% | | 84.19% | 85.07% | 0% | 84.65% | | 86.30% |

DNN model could be retrained to maintain high accuracy. Accompanied by accuracy recovery, updated weights on chip will vary from chip to chip. The transferability of the adversarial examples are strongly suppressed by the finetune for each chip instance. While classification accuracy of original attacked chip drops to almost 0%, other chips with adversarial CIFAR-10 images could still maintain more than 62% and 85% accuracy for VGG-8 and DenseNet-40, respectively.

Table 2: C&W attack on VGG-8 with different distance matric

| | | Attack original | | Attack retrained chip | |
|---|---|---|---|---|---|
| | | Software attack | Attack on chip1 | Chip2 acc. after attack | Attack on chip1 |
| Acc. before attack | | 91.96% | 88.10% | 90.5% | 90.78% |
| C&W attack | $L_0$ attack | 0.57% | 73.54% | 0.261% | 71.3% |
| | $L_2$ attack | 0.68% | 74.23% | 0.024% | 63.4% |
| | $L_\infty$ attack | 2.61% | 73.35% | 0.879% | 70.1% |